\documentclass[letterpaper, 10 pt, conference]{ieeeconf}  

\IEEEoverridecommandlockouts                              

\usepackage[usenames, dvipsnames]{color}
\usepackage[dvipsnames]{xcolor}
\usepackage{amsmath}
\usepackage{mathtools}
\usepackage{amsfonts}
\usepackage{amssymb}
\usepackage{tikz}
\usetikzlibrary{positioning,shapes,arrows}
\usepackage{graphicx}
\usepackage[hyphens]{url}
\usepackage{hyperref}
\usepackage[caption=false]{subfig}
\usepackage{cite}

\usepackage{fancyhdr}
\fancypagestyle{specialfooter}{%
\fancyhf{}
\fancyfoot[L]{Some test text}
\fancyfoot[C]{\thepage}
}

\newenvironment{copyrightnoticeFont}{\fontsize{7pt}{8pt}\selectfont\fontfamily{phv}\selectfont}{\par}

\usepackage[final]{changes}
\definechangesauthor[name={Per cusse}, color=blue]{per}

\title{\bf Deep Neural Networks for Improved, Impromptu Trajectory Tracking of Quadrotors}

\author{Qiyang Li, Jingxing Qian, Zining Zhu, Xuchan Bao, Mohamed K. Helwa, and Angela P. Schoellig\thanks{The authors are with the Dynamic Systems Lab (www.dynsyslab.org) at the University of Toronto Institute for Aerospace Studies (UTIAS), Canada. Email: \{qiyang.li, jingxing.qian, zining.zhu, xuchan.bao\}@mail.utoronto.ca, mohamed.helwa@robotics.utias.utoronto.ca, schoellig@utias.utoronto.ca.}}

\begin{document}

\maketitle
\thispagestyle{fancyplain}
\renewcommand{\headrulewidth}{0pt}
\pagenumbering{gobble}
\lfoot{\begin{copyrightnoticeFont}\vspace{-2em}
\textbf{Accepted final version}. To appear in \textit{the 2017 IEEE International Conference on Robotics and Automation}.\\
\copyright2017 IEEE. Personal use of this material is permitted. Permission from IEEE must be obtained for all other uses, in any current or future media, including reprinting/republishing this material for advertising or promotional purposes, creating new collective works, for resale or redistribution to servers or lists, or reuse of any copyrighted component of this work in other works.\end{copyrightnoticeFont}}
\pagestyle{empty}
\begin{abstract}
Trajectory tracking control for quadrotors is important for applications ranging from surveying and inspection, to film making. However, designing and tuning classical controllers, such as  proportional-integral-derivative (PID) controllers, to achieve high tracking precision can be time-consuming and difficult, due to hidden dynamics and other non-idealities. 
The Deep Neural Network (DNN), with its superior capability of approximating abstract, nonlinear functions, proposes a novel approach for enhancing trajectory tracking control.
This paper presents a DNN-based algorithm \added[id=per, remark={\dotsc}]{as an add-on module} that improves the tracking performance of a classical feedback controller. Given a desired trajectory, the DNNs provide a tailored \added[]{reference} input to the controller based on their gained experience. The input aims to achieve a unity map between the desired and the output trajectory.
The motivation for this work is an interactive ``fly-as-you-draw" application, in which a user draws a trajectory on a mobile device, and a quadrotor instantly flies that trajectory with the DNN-enhanced control system. 
Experimental results demonstrate that the proposed approach improves the tracking precision for  user-drawn trajectories\ after the DNNs are trained on selected periodic trajectories, suggesting the method's potential in real-world applications. Tracking errors are reduced by around 40-50\% for both training and testing trajectories from users, highlighting the DNNs' capability of generalizing knowledge.

\end{abstract}

\section{Introduction}
\label{sec:intro}
In recent years, quadrotors have been widely used for   civilian and law-enforcement  purposes, such as   providing aerial surveillance, carrying out rescue missions, transporting goods over distance, and performing surveying and inspection tasks~\cite{drug_traffickers, drone_rescue, drone_delivery, michael2011cooperative}. In all these applications, the quadrotor is required to  \added[]{precisely track a desired trajectory}\deleted[]{track a desired trajectory or spatial path,} in order to perform the task safely and effectively. 


Trajectory tracking for quadrotors poses a challenge on controller design. First, quadrotors are underactuated systems with nonlinear dynamics, making it a difficult control problem. Second, trajectory tracking precision of quadrotors can be affected by many factors, including uncertainty in the turn-rate-to-thrust map, time delays that are difficult to quantify, aerodynamic effects and other unpredictable factors such as  friction in the actuators. \deleted[]{However}\added[]{Third}, even in a perfect world, where the system dynamics are known exactly, a given classical controller cannot achieve perfect tracking for any arbitrary, \deleted[]{generally }feasible, desired trajectory. 

\begin{figure}[t]
    \centering
    \includegraphics[width=8.5cm]{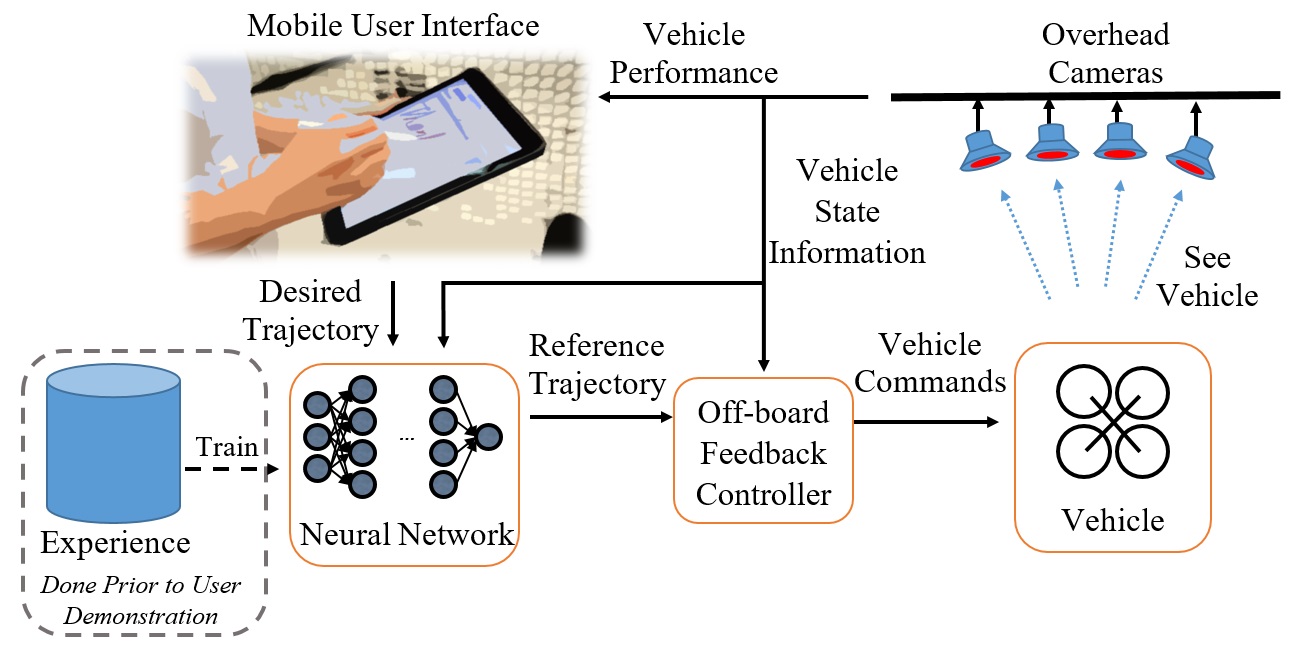}
    \caption[caption]{Block diagram of our interactive \added[]{``fly-as-you-draw''} demo. The user draws an arbitrary trajectory on a mobile device. Upon receiving the new trajectory, the quadrotor immediately takes off and follows the signal, processed by the pre-trained Deep Neural Network (DNN) in the ($x$-$z$)-plane. The overhead camera system provides state feedback and also performance feedback to the user. \added[]{A demo video can be found at \\ \url{http://tiny.cc/DNN-ImpromptuTracking}.}}
    \label{fig:cover}
\end{figure}

Our goal is to achieve improved trajectory tracking control for quadrotors while taking into account three features that are crucial for most real-world trajectory tracking applications (Figure \ref{fig:cover} shows our specific application): 
\begin{enumerate}
    \item Stability of the control system and robustness to reasonable disturbances must be guaranteed to ensure safety of the operation.
    \item \deleted[id=per, remark={}]{The tracking control system must be independent of a specific trajectory.} \added[id=per, remark={}]{\deleted[id=per,remark={}]{in other words, t}The system should be able to precisely track a new trajectory without adaptation.} 
    \item The computational resources needed for the control system should be manageable such that the algorithm can be applied to small vehicles with limited computational power.
\end{enumerate}

Simple controllers such as typical proportional-integral-derivative (PID) controllers can achieve adequate performance under certain conditions, for example low speeds and accelerations, while having all the crucial features mentioned above~\cite{5569026, Lupashin201441}. However, PID controllers are difficult to tune and they tend to behave poorly on more aggressive trajectories. There exist\deleted[]{s} previous works on improving control for quadrotors or other robots, such as learning the dynamics or the inverse dynamics, iterative learning control and Gaussian Process learning. \added[]{However,} we show in Section~\ref{sec:RelatedWork} that these approaches have drawbacks\deleted[]{, however,} with respect to the three crucial features we identified above, which are relevant for real-time trajectory tracking.



In this paper, we propose a DNN-based \added[]{control system} which improves the trajectory tracking performance by utilizing past flight experiences. After offline training from relevant flight examples, a generalized model is obtained with the DNN. 
This model can be evaluated in real-time\deleted[]{,} to modify the \deleted[]{control}\added[]{reference} signal given to the controller\deleted[]{as the reference signal}. With no prior knowledge of the system other than the training data, the proposed method demonstrates its ability to reduce trajectory tracking error by compensating for controller imperfections and unknown dynamics. Also, the DNN model is computationally efficient for real-time evaluation and effective even on \deleted[]{arbitrary,}\added[]{arbitrary trajectories, not trained on before,}\deleted[]{non-trained trajectories}\deleted[]{not trained on before,} making it applicable to impromptu tracking tasks.

\begin{figure}[t]
    \centering
    \vspace{3mm}
    \includegraphics[width=8.2cm]{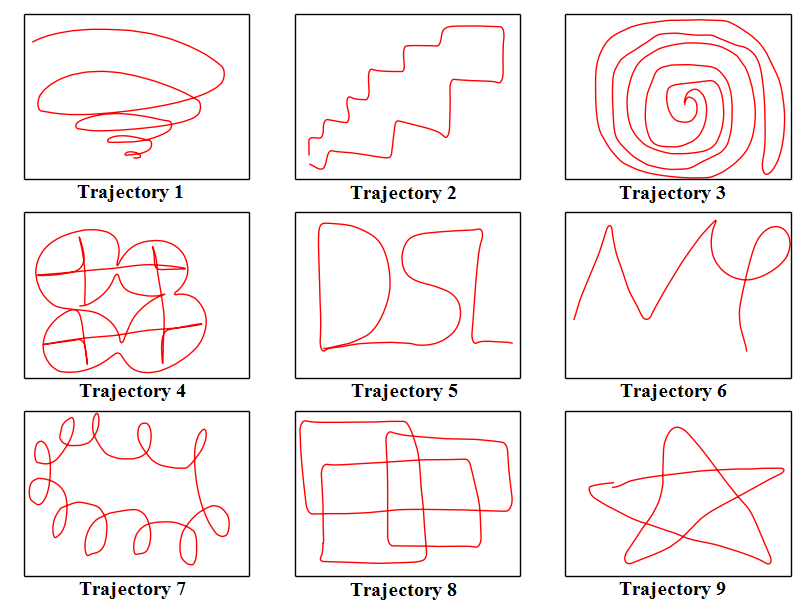}
    \caption{Nine of the 30~user-drawn trajectories used in testing the performance of the DNN control system. \added[]{For testing on the quadrotors, the drawn trajectories were modified to ensure the trajectory is bounded by maximum velocity (0.6$~\text{m/s}$) and acceleration (2.0$~\text{m/s}^2$).}}
    \label{fig:9drawn}
\end{figure}

To validate the effectiveness of the proposed method and motivate this work, we implement an interactive fly-as-you-draw application, where the quadrotor takes off to follow an arbitrary, hand-drawn trajectory immediately after the user finishes drawing the trajectory. The application uses neural networks, pre-trained with quadrotor flight data collected from periodic training trajectories, to obtain reference signals in real-time for an off-board feedback controller. This process is described in Figure \ref{fig:cover}. With this interactive application, we evaluate different DNN features, and compare the DNN performance with the baseline \deleted[]{PD-control }\added[]{nonlinear controller}\deleted[]{system} performance. Nine of the 30 user-drawn testing trajectories used in our experiments are shown in Figure \ref{fig:9drawn}. Through the experiments, we demonstrate that the proposed approach, with proper feature selection for the DNN learning, is able to consistently enhance trajectory tracking precision for complex, arbitrary hand-drawn trajectories. \deleted[]{This demonstrates its potential in real-world applications that require highly precise maneuvering, such as monitoring and inspection tasks, aerobatics, skywriting and airborne filming.}\added[id=per, remark = {mention generalizability in Intro}]{Moreover, because the DNN serves as a pre-block outside the feedback control loop, the proposed method can be generalized as an add-on to any black-box, stable feedback control system. \deleted[]{which shows its potential in several control system applications.}}\added[]{These characteristics of the proposed approach demonstrate its potential in real-world applications that require highly precise maneuvering, such as monitoring and inspection tasks, aerobatics, skywriting and airborne filming.}

The paper is organized as follows. Section \ref{sec:RelatedWork} summarizes related previous work on advanced trajectory tracking control. In Section \ref{sec:probstatement}, we state the problem, followed by the general methodology in Section \ref{sec:method}. The experimental setup is presented in Section \ref{sec:experiment}, and the corresponding results are presented in Section \ref{sec:experimentalResults}. A brief summary and discussion of the results are presented in Section \ref{sec:conclusion}.

\section{Related Work}
\label{sec:RelatedWork}

Neural networks (NNs) are a generic approach for approximating functions given a large amount of data.
\added[]{In previous work, NNs have been adopted to introduce modifications to feedback control loops.}\deleted[]{Previous work has demonstrated that NNs are able to learn the dynamics and the inverse dynamics of unmanned aerial vehicles (UAVs)~\cite{i_2015_DL_models, mohajerin2014modular, a_2014_DIC}.}
The papers~\cite{i_2015_DL_models} and~\cite{mohajerin2014modular} use NNs to learn the dynamics of a helicopter and a quadrotor\added[]{,} respectively. In~\cite{a_2014_DIC}, an NN is used for direct inverse control of a quadrotor with promising initial simulated results for hover flight. \deleted[]{Both learning the dynamics and the inverse dynamics involve modification of the original feedback control loop.}\added[]{However, by involving NNs to modify the original feedback control loop, the stability of the control system will likely be affected.}\deleted[]{One of the biggest challenges of that is to ensure stability of the control system.} When previously unseen inputs are given to the NN \added[id=per, remark={instability}]{that is part of the feedback control loop}, the NN might generate unpredictable outputs, leading to instability of the system. Instead, we use DNNs (multi-layer NNs) to learn a model that directly determines the reference inputs to the feedback control loop. The proposed DNNs act as a pre-block outside the original feedback control loop, and run at a lower update rate, which makes the system much less susceptible to instability.

Iterative learning control (ILC) is an approach of improving the control precision by repeating the same task and learning from previous executions~\cite{Schoellig2012}. Through the repetition of one specific task, ILC learns an updated reference input and achieves high-precision tracking for this particular task. Unlike simple controllers that can fail to achieve aggressive maneuvers, previous work has demonstrated ILC's ability of achieving high control precision on these tasks~\cite{mueller2012iterative}. One significant drawback of this approach is that the experience of learning one specific task is not transferable to other tasks. Although \cite{2013_ILC_transfer} has shown that linear maps can optimize ILC initialization from previous experience, ILC still has to re-learn through multiple iterations before achieving high precision for a new trajectory. Our approach allows training ahead of time, and the trained model generalizes to arbitrary trajectories without any adaption process. This feature makes it suitable for applications that require the vehicle to complete the desired task with high precision in a timely manner.  

Gaussian Process (GP) learning is receiving growing attention in the control community and has been used in various control problems. For instance, an accurate kinematic control of a cable-driven surgical robot is implemented in \cite{2014_surgical}. Similar to the idea of learning the reference input in our proposed method, the GP learns the reference input to the controller of the surgical robot, improving the tracking precision of the end-effector. Our approach is different from this GP learning approach in two ways: 1) we apply the method on the quadrotor system which has different dynamics compared to surgical robots; 2) we employ DNNs as our learning technique instead of GPs. One advantage of using DNNs is that DNNs can summarize data using a fixed-size model. A GP model gets bigger as more data is collected, making the model large in terms of required storage and computationally expensive to evaluate. In contrast, when the size of the data set increases, DNNs adjust their parameters to better fit the training data without increasing the model size. Since modeling complex relations usually requires a large set of training data, the invariant model size makes the DNNs more promising on control systems with complex dynamics\deleted[]{and}, especially\deleted[]{,} when computation is limited.

\section{Problem Statement}
\label{sec:probstatement}
For a given dynamic system with a baseline feedback controller (see Figure \ref{fig:control_loop}), the problem is to learn a mapping from the desired trajectory $T_d$ and the current state $\mathbf{s}_c$ to the reference input $\mathbf{s}_r$ of the baseline controller, in order to enhance the tracking performance of the overall system for arbitrary desired trajectories.  We define the desired trajectory $T_d = \{\mathbf{s}_{d,1}, \mathbf{s}_{d,2}, \dotsc, \mathbf{s}_{d,N}\}$ as a sampled trajectory containing $N$ consecutive time steps, where $\mathbf{s}_{d,t}$ represents the desired state at the $t^{\text{th}}$ time step. Learning \added[]{is}\deleted[]{should be} done off-line, and the learned mapping is applied in real-time.

\section{Methodology}
\label{sec:method}

\deleted[id=per,remark = {}]{Our DNN learning approach aims to address the problem stated in Section \ref{sec:probstatement}. In this section, we first introduce supervised learning with DNNs as the foundation in Section \ref{sec:learningDNN}. Based on this, the proposed control design is presented in Section \ref{sec:controldesign}. The last subsection, Section \ref{sec:featureselection}, presents the importance of feature selection for DNN learning. }

\tikzstyle{block} = [draw, fill=blue!20, rectangle, 
    minimum height=2em, minimum width=4em]
\tikzstyle{sum} = [draw, fill=blue!20, circle, node distance=1.5cm]
\tikzstyle{input} = [coordinate]
\tikzstyle{output} = [coordinate]
\tikzstyle{pinstyle} = [pin edge={to-,thin,black}]
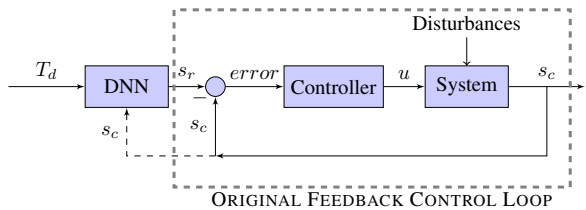
\begin{figure}[t]   
\centering
   \vspace{3mm}
    \resizebox{.9\hsize}{!}{
        \centering
        \begin{tikzpicture}[auto, node distance=2cm,>=latex']
            \node [input, name=input] {};
            \node [block, right of=input] (DNN) {DNN};
            \node [sum, right of=DNN] (sum2) {};
            \node [block, right of=sum2] (controller) {Controller};
            \node [block, right of=controller, pin={[pinstyle]above:Disturbances},
                    node distance=2.25cm] (system) {System};
            \draw [->] (controller) -- node[name=u] {$u$} (system);
            \node [output, right of=system] (output) {};
            \node [input, below=1cm of sum2, name = measurements] {Measurements};
        
            \draw [draw,->] (input) -- node {$T_d$} (DNN);
            \draw [->] (DNN) -- node {$s_r$} (sum2);
            \draw [->] (sum2) -- node {$error$} (controller);
            \draw [->] (system) -- node [name=y] {$s_c$}(output);
            \draw [->] (y) |- (measurements);
            \draw [dashed, ->] (measurements) -| node[pos=0.99] {$ $} 
                node [near end] {$s_c$} (DNN);
            \draw [->] (measurements) -| node[pos=0.99] {$-$} 
                node [near end] {$s_c$} (sum2);
            
                \draw [color=gray,line width=0.5mm,dashed](9.5,-1.7) rectangle (2.8,1.3);
                \node at (3.3,-1.9) [above=5mm, right=0mm] {\textsc{Original Feedback Control Loop}};
        \end{tikzpicture}}
    \caption{The modified control system with the DNN block in front of the original baseline controller\deleted[]{where}\added[]{;} the controller takes in the reference state (produced by the DNN) and the current state to control the system.}
   \label{fig:control_loop}
\end{figure}

\subsection{Supervised Learning with Deep Neural Network Model}
\label{sec:learningDNN}
\added[]{Our approach builds upon supervised learning with DNN. This learning process requires the preparation of a large number of labeled training examples and the training of DNN on these examples. Each labeled training example consists of an input and expected output pair to describe what the function should output according to a specific input. The training of DNN \deleted[]{on}\deleted[]{the label training}\deleted[]{these examples} involves back-propagation to minimize the loss over all training examples~\cite{lecun2015deep}, defined as the Euclidean distance between the network's output and expected output. After learning from the labeled training examples, the DNN can summarize a mapping from the training inputs to the training outputs.}

\added[]{A feed-forward DNN with \added[]{rectified linear}\deleted[]{ReLU activation} units \added[]{(ReLU)} is used to learn the mapping formulated in Section \added[]{\ref{sec:probstatement}}\deleted[]{\ref{sec:method}}. To prepare the training examples for learning the target mapping from the flying data, we use the actual trajectory as training inputs and the reference signal as the labeled output. \added[]{The idea behind this selection is that if} the actual trajectory was the desired trajectory, then the DNN should provide \added[]{this saved}\deleted[]{the} reference signal to achieve perfect tracking. The specification of input and outputs for the DNN will be discussed in\deleted[]{greater} detail in Section \ref{subsec:data}.}




\subsection{DNN as Reference Generator}

\label{sec:controldesign}

\added[]{The proposed method modifies the control system design}\deleted[]{We modify the control system} by adding a DNN block in front of the controller. At each time step $t$, the trained DNN modifies the control signal in real time by giving the reference input $\mathbf{s}_{r,t}$ to the controller based on the\deleted[]{required} desired trajectory $T_d$ as well as the current state of the quadrotor $\mathbf{s}_{c,t}$. 

Figure \ref{fig:control_loop} highlights the difference between the original control system and the proposed one. The reference states generated by the DNN over $N$ consecutive time steps form the reference trajectory $T_r = \{\mathbf{s}_{r,1}, \mathbf{s}_{r,2},\dotsc, \mathbf{s}_{r,N}\}$, where $\mathbf{s}_{r,t}$ is the reference state generated by the DNN at the $t^{\text{th}}$ time step. Also, the actual trajectory $T_c = \{\mathbf{s}_{c,1}, \mathbf{s}_{c,2}, \dotsc,\mathbf{s}_{c,N}\}$ completed by the vehicle over the $N$ consecutive time steps is observed. The actual trajectory $T_c$ is expected to closely match the desired trajectory $T_d$. 


In control systems, current state feedback enables the control system to reject external disturbances if the controller is designed properly. Similarly, the extra loop introduced in our proposed system enables the DNN to adjust its output reference according to the current state to compensate the disturbances. 
We choose the feedback rate for this extra loop to be much lower than the original control loop to ensure that the stability of the original control system is not disrupted by the DNN signals.
For example, in our experiments on the quadrotor control system, we design our DNN to send reference states at 7 Hz, which is 10 times slower than the control loop operating at 70 Hz. Therefore, DNN control signals can be non-intrusive to the original control system.

\subsection{Feature Selection for the DNN}
\label{sec:featureselection}
Ideally, at each time step $t$, the DNN would receive all the given information (both the entire desired trajectory $T_d$ and the current state $\mathbf{s}_{c,t}$), and produce an optimal reference state $\mathbf{s}_{r,t}$ as the input to the controller to minimize the quadrotor's tracking error. 
However, this makes input dimension huge, and requires exponentially increasing amount of training data. Therefore, selecting proper state information for DNN is crucial for making the DNN learning effective.

\added[]{A minimum feature selection is to only use the current desired state $\mathbf{s}_{d,t}$ and the current state $\mathbf{s}_{c,t}$ as the input, but this configuration may not be able to provide information for the DNN to model the hidden dynamics including time delay which may deteriorate the tracking performance. Hence, we consider including future desired states into the \added[]{DNN }input. 
With the additional future information, the properly-trained DNN is expected to plan the control ahead of time by considering future desired states and improve control performance. 
In this paper, we investigate the influence of future states on DNN learning. It is hypothesized that the DNN can give a better performance when the desired states in the near future are introduced into the DNN input.}
\deleted[]{Real-world complex control systems usually have hidden dynamics including time delay, which deteriorates the tracking performance and may even cause instability problems. Hence, planning the control ahead of time by considering future desired states is expected to improve the control performance. Model Predictive Control (MPC), as one of the advanced control techniques that takes future desired states into account, can optimize a proper control sequence ahead of time based on a dynamic model of the system~\cite{camacho2013model}. Instead of optimizing the control sequence based on a pre-determined model, we expect that properly-trained DNN can directly determine an optimized control signal based on the future desired states and the current state of the system. We hypothesize that the DNN considering the desired states in the near future can give a better performance.}


To validate this hypothesis, we conducted experiments on a quadrotor in real-world environment to investigate 1) the effect of selecting different state information, including future desired states and current state feedback, as the input of the DNN on the DNN performance, and 2) the generalizability of this method for improving the tracking performance for different trajectories overall.









\section{Experiment Setup}
\label{sec:experiment}

\subsection{\added[]{The Quadrotor Model and }Experiment Platform}
\added[]{This subsection provides a glimpse of quadrotor dynamics as well as the experiment platform. For more details about the quadrotor dynamics and control, readers are referred to~\cite{5569026}.}

\label{sec:quadrotor}
\added[]{A typical quadrotor consists of a symmetrical cross-shaped frame with four propellers mounted at the end of four arms. \added[]{The full state of the quadrotor consists of 12 components}. The translational position of the quadrotor's center of mass is defined as $\mathbf{p} = (x, y, z)$ and the attitude, represented by Euler angles roll, pitch and yaw, is defined as \(\left( \phi,\theta,\psi \right) \). In addition to translational position and attitude, the full state\deleted[]{, $\mathbf{s}_c$,} of the quadrotor includes the translational velocity,   \(\mathbf{v} = \left( \dot{x},\dot{y},\dot{z} \right) \), and the rotational velocity, \(\boldsymbol{\omega}  = \left( p,q,r \right)\). \deleted[]{The acceleration in the $z$-axis, $\ddot{z}$, is also included because the controller we used in the experiment requires $\ddot{z}$ along with the other $12$ state components as its inputs. In summary, the full state of the quadrotor is defined as $\mathbf{s}_c = \{x,y,z,\dot{x}, \dot{y}, \dot{z}, \phi, \theta, \psi, p, q, r, \ddot{z}\}$.}}

The experiments are conducted on a Parrot AR.Drone 2.0 quadrotor. This commercial quadrotor suits the needs of this study as it features highly nonlinear dynamics, complex aerodynamics that are hard to model, and most importantly, an unmodified black-box, which is an on-board controller that controls the vehicle's roll, pitch and yaw by adjusting motor forces. 
The quadrotor's states are all measured by the overhead \emph{Vicon} motion capture system. The system features eight 4-mega pixel Vicon cameras running at 200 Hz. A similar experimental setup is described in detail in~\cite{Lupashin201441}.
The baseline control system used in this paper consists of two controllers: the on-board controller and the off-board controller.
\deleted[]{The off-board controller is a standard PD controller implemented using the open-source Robot Operating System (\emph{ROS}). The off-board controller, which runs at 70 Hz, receives the quadrotor's current state and the desired state, and outputs to the on-board controller the desired roll ($\phi_{cmd}$), pitch ($\theta_{cmd}$), one of the three elements of rotational velocity ($r_{cmd}$), and velocity in the $z$-direction ($\dot{z}_{cmd}$). }
\added[]{The off-board controller is a nonlinear controller, composed of a nonlinear transformation and standard PD controller. It is implemented using the open-source Robot Operating System (\emph{ROS}). The controller runs at 70 Hz, \deleted[]{which} receives the quadrotor's current state and the \deleted[]{desired state}\added[]{reference}, and outputs to the on-board controller the desired roll, pitch, yaw velocity and $z$ velocity ($ \phi_{cmd}$, $\theta_{cmd}$, $r_{cmd}$, $\dot{z}_{cmd}$).}
The on-board controller runs at 200 Hz, receives the four commands from the off-board controller, and adjusts the four motor thrusts $F_1, \dotsc,F_4$ accordingly. The DNN feedback loop runs at 7 Hz, which is 10 times slower than the off-board controller.

\subsection{Task Performance}
Each task performed by the quadrotor involves following one of the pre-defined\added[]{,} desired trajectories $T_d$ in the ($x$-$z$)-plane, where these trajectories are hand drawn through our interactive application (Figure \ref{fig:cover}). 
The error function for each task is defined as the root-mean-square (RMS) error  of \emph{N} pairs of $(x,y,z)$-coordinates sampled at 7 Hz, the DNN feedback loop sampling rate, between the desired trajectory, $T_d$, and the observed trajectory, $T_c$: 
\begin{equation}
\label{eq_rms}
    E(T_{c}, T_{d}) = \sqrt{\frac{1}{N} \sum_{t=1}^{N} \|\mathbf{p}_{c,t} - \mathbf{p}_{d,t}\|^2},
\end{equation} 
where $\|\mathbf{p}_{c,t} - \mathbf{p}_{d,t}\|$ is the Euclidean norm, while $\mathbf{p}_{d,t}$ and $\mathbf{p}_{c,t}$ are the position coordinates sampled at the $t^{\text{th}}$ time step from the desired trajectory $T_d$ and the observed trajectory $T_c$, respectively.
The quadrotor in the experiment repeats each task with and without the aid of the \deleted[]{same set of trained DNNs}\added[]{trained DNN}. The percentage reduction in errors between corresponding flights is identified as the improvement of our method on this specific task:
\begin{equation}
    I(T_{\text{w/ DNN}}, T_{\text{w/o DNN}}) = \Big(1 - \frac{E_{\text{w/ DNN}}}{E_{\text{w/o DNN}}}\Big) \times 100\%,
\end{equation}
where $E_{\text{w/ DNN}}$ and $E_{\text{w/o DNN}}$ are the RMS errors in \eqref{eq_rms} with and without the DNN, respectively.

\subsection{DNN Input-Output Specification}
\label{subsec:spec}

In general, the trained DNN provide a mapping from the current and selected desired states to the reference state:
\begin{equation}
    \{\mathbf{s}_{c,t}, \mathbf{s}_{d,t_1}, \dotsc, \mathbf{s}_{d,t_L}\} \rightarrow \mathbf{s}_{r, t},
\label{eq:map}
\end{equation}
where $\mathbf{s}_{c,t}$ is the vehicle's current state at the $t^{\text{th}}$ time step, and $\{\mathbf{s}_{d,t_1}, \dotsc,\mathbf{s}_{d,t_L}\}$ are $L$ selected desired states from the desired trajectory $T_d$. \added[]{Each of the states ($\mathbf{s}_c$, $\mathbf{s}_d$ and $\mathbf{s}_r$) mentioned above contains the full state of the quadrotor along with the translational acceleration on $z$-direction, $\ddot{z}$. Among all translational accelerations, $\ddot{x}$, $\ddot{y}$ and $\ddot{z}$, only $\ddot{z}$ is included in these states because the controller we used in the experiment only requires $\ddot{z}$ along with the full state of the quadrotor, $\{x,y,z,\dot{x}, \dot{y}, \dot{z}, \phi, \theta, \psi, p, q, r\}$, as its inputs. In summary, the state of the vehicle is defined as $\mathbf{s} = \{x,y,z,\dot{x}, \dot{y}, \dot{z}, \phi, \theta, \psi, p, q, r, \ddot{z}\}$.}

Based on this general input-output mapping provided by the DNN, we explore three different configurations of the DNN to investigate the influence of including the future desired states and/or the current state feedback as inputs to the DNN (as discussed in Section \ref{sec:featureselection}):
\begin{itemize}
    \item DNN with future desired states and the current state feedback;
    \item DNN with future desired states and without the current state feedback; 
    \item DNN without future desired states and with the current state feedback.
\end{itemize}

All configurations consider one or more desired states at different time steps over the entire flight path as part of the input. 
Since we hypothesize that using future desired states can enhance tracking performance, we focus on the scenario where we only select the desired states $\{\mathbf{s}_{d,t_1}, \dotsc,\mathbf{s}_{d,t_L}\}$ from the current desired state and the future desired states, i.e., we select $t_i=t+\Delta_i$, where $\Delta_i \geq 0$, for $i=1,\dotsc,L$.  
 
For the first configuration with the current state feedback, the current observed state $\mathbf{s}_{c,t}$ and the $L$ selected desired states, $\{\mathbf{s}_{d,t+\Delta_1}, \dotsc,\mathbf{s}_{d,t+\Delta_L}\}$, are given to the DNN at each time step. The actual input to the DNN consists of two major parts. The first part includes $L + 1$ sets of $\{\dot{x}, \dot{y}, \dot{z}, \phi, \theta, \psi, p, q, r, \ddot{z}\}$ 
\deleted[id=per,remark=obsolete]{\footnote{The elements $\dot{x}, \dot{y}, \dot{z}, \phi, \theta, \psi, p, q, r$ all come from the state of the quadrotor defined in Section \ref{sec:quadrotor}, while $\ddot{z}$ is the extra variable given to the DNN to assist the training.}}  from the current observed and the selected desired states. The second part includes 
$L$ sets of desired positions relative to the current observed position
$\{\mathbf{p}_{d,t+\Delta_1} - \mathbf{p}_{c,t}, \dotsc,\mathbf{p}_{d,t+\Delta_L} - \mathbf{p}_{c,t}\}$, where $\{\mathbf{p}_{d,t+\Delta_i}\}$ and $\mathbf{p}_{c,t}$ are the position components from the selected desired states and the current observed state, respectively\footnote{Relative position information, instead of absolute position information, is used in order to reduce input data dimension.}. The input to the DNN for the second configuration is similar to the first configuration with only $\mathbf{s}_{c,t}$ being replaced by $\mathbf{s}_{d,t-1}$, and consequently $\mathbf{p}_{c,t}$ being replaced by $\mathbf{p}_{d,t-1}$, where $\mathbf{p}_{d,t-1}$ is the position components from $\mathbf{s}_{d,t-1}$. 
The last configuration is a special case of the first configuration, in which the current desired state $\mathbf{s}_{d,t}$ is the only selected desired state in the DNN input ($L = 1$ and $\Delta_1 = 0$). For this configuration, the DNN does not consider the future states, and generates the reference state only based on the current observed state and the current desired state.


The output of the DNN in the three configurations is the reference state $\mathbf{s}_{r,t}$. In our experiment, we reduce data complexity by only learning the difference between the reference state $\mathbf{s}_{r,t}$ and the current desired state $\mathbf{s}_{d,t}$ in translational position and velocity components. 

\subsection{Data Collection for DNN Training}
\label{subsec:data}
To train the DNN, we need to collect the state data from real-world flights and select the training data.
To that end, we design a 400-second trajectory that oscillates sinusoidally in all the $x$, $y$ and $z$-directions with different combinations of \deleted[]{frequencies and} amplitudes to cover the feasible state space as much as possible. In particular, each of the three directions has its own oscillating frequency (0.27~Hz, 0.20~Hz and 0.13~Hz for the $x$, $y$ and $z$-directions respectively). We also gradually increase the amplitudes from 0 to 2~m in all directions. On this trajectory, the quadrotor can reach a maximum velocity of 1.5 m/s and a maximum acceleration of 4 m/s$^2$. The maxima for rotational velocity $\boldsymbol{\omega} $ and rotational acceleration $\dot{\boldsymbol{\omega} }$ are 0.4 rad/s and 1 rad/s$^2$ respectively.

Using the baseline control system to follow the designed training trajectory, we collect a  $\{\mathbf{s}^*_{c,t}$, $\mathbf{s}^*_{d,t}\}$ pair at each time step, where $\mathbf{s}^*_{c,t}$ is the current observed state and $\mathbf{s}^*_{d,t}$ is the current desired state\footnote{The superscript * indicates a state from the training data.}. Approximately 10,000 raw data pairs are collected at a 7 Hz rate from four flights on the training trajectory, and then organized in the consecutive time order.



We select and re-organize these raw data pairs to establish the labeled training set.
Recall that the DNNs aim to learn a mapping from the current state $\mathbf{s}_{c,t}$ and L selected desired states $\{\mathbf{s}_{d,t+\Delta_1}, \dotsc, \mathbf{s}_{d,t+\Delta_L}\}$ to the reference state $\mathbf{s}_{r,t}$ that should be given to the controller at the current time step. For any pair $\{\mathbf{s}^*_{c,t}$, $\mathbf{s}^*_{d,t}\}$ in the data set, we consider $\{\mathbf{s}^*_{c,t+\Delta_1\added[]{+1}}, \dotsc,\mathbf{s}^*_{c,t+\Delta_L\added[]{+1}}\}$. If we treat $\mathbf{s}^*_{c,t}$ as the current observed state $\mathbf{s}_{c,t}$ of the quadrotor and $\{\mathbf{s}^*_{c, t+\Delta_1\added[]{+1}}, \dotsc, \mathbf{s}^*_{c, t+\Delta_L\added[]{+1}}\}$ as the selected desired states $\{\mathbf{s}_{d, t+\Delta_1}, \dotsc, \mathbf{s}_{d, t+\Delta_L}\}$, \deleted[]{then $\mathbf{s}^*_{d,t}$ is a feasible solution of the reference state $\mathbf{s}_{r,t}$}\added[]{then $s^*_{d,t}$ may be selected as the reference state $s_{r,t}$, given that $s^*_{d,t}$ is a point in a feasible reference sequence achieving perfect tracking with one sample delay}. The mapping $\{\mathbf{s}^*_{c,t}, \mathbf{s}^*_{c,t+\Delta_1\added[]{+1}}, \dotsc, \mathbf{s}^*_{c,t+\Delta_L\added[]{+1}}\} \rightarrow \mathbf{s}^*_{d,t}$, therefore, is an approximate mapping of $\{\mathbf{s}_{c,t}, \mathbf{s}_{d,t+\Delta_1}, \dotsc, \mathbf{s}_{d,t+\Delta_L}\} \rightarrow \mathbf{s}_{r,t}$.
Thus, for each pair $\{\mathbf{s}^*_{c,t}$, $\mathbf{s}^*_{d,t}\}$ in the data set, we can form a training pair ($a$, $b$) for learning the approximate mapping, where $a = \{\mathbf{s}^*_{c,t}$, $\mathbf{s}^*_{c,t+\Delta_1\added[]{+1}}, \dotsc, \mathbf{s}^*_{c,t+\Delta_L\added[]{+1}}\}$ is the input and $b = \mathbf{s}^*_{d,t}$ is the labeled output. A labeled data set can then be obtained by collecting the training pairs ($a$, $b$). As a result of training, using the supervised learning technique for our DNNs with the labeled data (as discussed in Section \ref{sec:learningDNN}), the DNNs are expected to learn an approximate mapping from the inputs to the output.

\subsection{DNN Training}
\label{subsec:train}
We train six different DNNs \added[id=per, remark={\dotsc}]{that share the same input states} to find the mapping in \eqref{eq:map}, one for each of the 
\added[]{position and velocity elements of the reference state $\mathbf{s}_r$, $\{x,y,z, \dot{x},\dot{y},\dot{z}$\}. Other elements in $\mathbf{s}_r$, \added[]{$\{\phi, \theta, \psi, p, q, r, \ddot{z}\}$}, maintain intact.} \added[]{We construct the DNNs in such a way because in preliminary experiments we observed that training of the six outputs might require different number of iterations to converge. For example, velocity and position in the $z$-axis are easier to train whereas positions in $x$-$y$ plane are much harder to train. We also observe that the six DNNs demonstrate comparable performance after 2,000 training iterations. However, it is possible to merge them to jointly learn \deleted[]{all}\added[]{the} six outputs. }  We construct the DNNs using \emph{TensorFlow}, an open-source library originally developed by Google. Each DNN consists of four fully connected hidden layers, and each layer contains 128 neurons. 90\% of the collected raw data pairs are used for training, while the rest are used for validation. Adam optimizer is used to tune the weight and bias parameters in the DNNs, and the learning rate is set at 0.0003~\cite{kingma2014adam}. To prevent over-fitting, a dropout rate of 0.5 is used~\cite{2014-dropout}. For each training iteration, 30 training pairs are used. 2,000 iterations are done for training each output\footnote{The 30 training pairs are randomly selected from the 90\% of the raw data pairs used for training.}.

\section{Experimental Results}
\label{sec:experimentalResults}

\subsection{Impact of Future States on DNN Tracking Performance}

\begin{figure}[t]
    \centering
    \includegraphics[height = 5cm,width=0.4\textwidth]{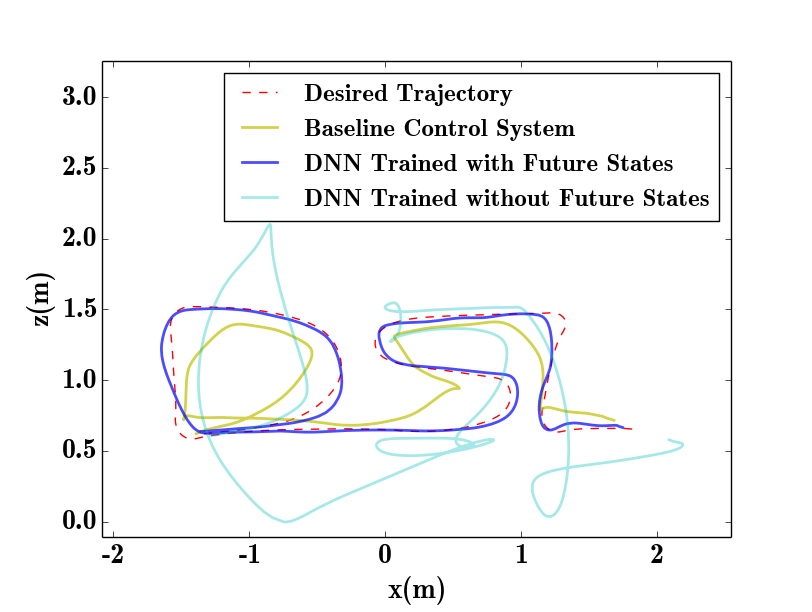}
        \caption{Observed trajectories resulting from DNNs with two future desired states ($\Delta_1 = 4$, +0.6~s, and  $\Delta_2 = 6$, +0.9~s, blue), DNNs without future desired states (cyan), and the baseline control system (yellow). The desired trajectory is Trajectory 5 in Figure \ref{fig:9drawn}. It consists of three hand-written letters ``DSL" (red dashed). The RMS tracking errors, \emph{E}, are 0.1926~m, 0.758~m and 0.434~m respectively. Overall, the proposed method with given future states achieved 56\% improvement over the baseline controller.}
    \label{fig:DNN_fvswf}
\end{figure}

\begin{figure}[t]
    \centering
        {\includegraphics[width=0.35\textwidth]{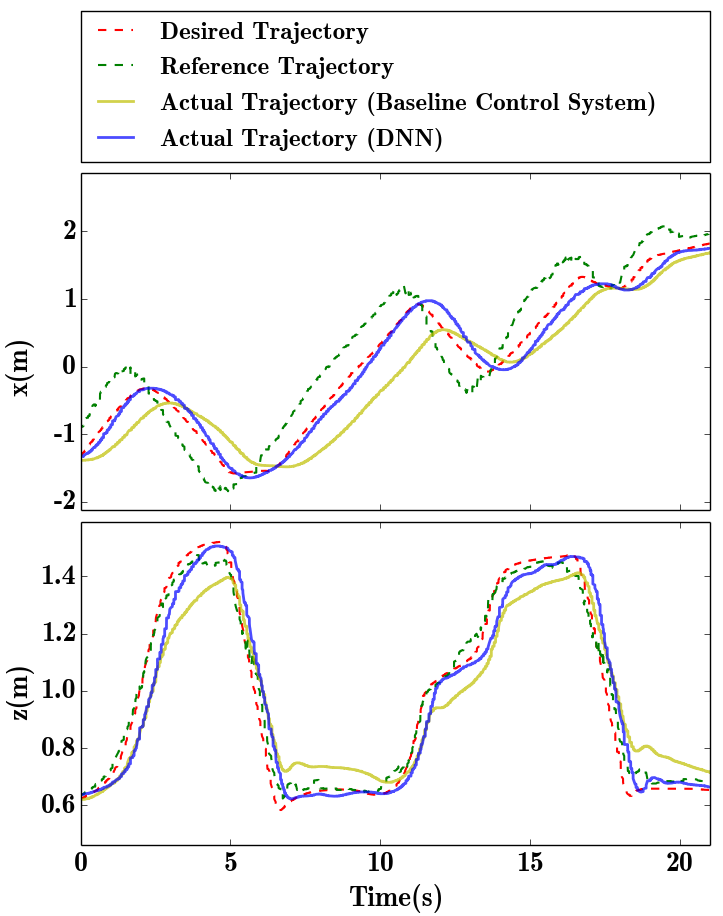}}
    \caption{\added[id=per, remark={\dotsc}]{The $x$ and $z$ performances of Figure \ref{fig:DNN_fvswf}. With the aid of the DNNs trained with future desired states, tracking error is significantly reduced in both axes.  Note that the green dashed line shows the reference trajectory calculated by the DNNs. The graph of $y$ performance is not shown since the $y$-components of velocity and position in the desired trajectory are zero. The RMS error for the proposed control system is 0.1481~m ($x$-axis), 0.0905~m ($y$-axis) and 0.0833~m ($z$-axis). The RMS error for the baseline control system is 0.401~m ($x$-axis), 0.1163~m ($y$-axis) and 0.1189~m ($z$-axis). The percentage improvement on each axis is 63\% ($x$-axis), 22\% ($y$-axis), 30\% ($z$-axis).}}
    \label{fig:DNN_fvswfe}
\end{figure}

The influence of introducing future desired states as part of the input to the DNNs is investigated. It is expected that the DNNs can use the future desired states information to better ``plan'' the flying path for the future desired states while compensating for the effect of hidden dynamics, including time-delay and other factors. In Figure \ref{fig:DNN_fvswf}, we show that the DNNs with future desired states and current state feedback performs significantly better than the baseline control and the DNNs with current state feedback but without future desired states. Also, Figure \ref{fig:DNN_fvswfe} highlights that the reference trajectories produced by the DNNs trained with future desired states and the current state feedback effectively correct the tracking error in both the $x$- and $z$-directions. 



\subsection{Impact of Feedback on DNN Tracking Performance}

We also investigate the influence of removing the current state feedback from the DNN inputs. Our experiments are conducted on the DNNs trained with future desired states. To remove the feedback loop, we keep the DNNs while replacing the current state feedback with the desired state from the previous time step during actual flight, as discussed in Section \ref{subsec:spec}. Table \ref{table:fbvsnfb} highlights that the \deleted[]{control system}\added[]{DNN} with current state feedback performs better than the \deleted[]{control system}\added[]{DNN} without current state feedback, while both obtain considerable improvements over the baseline control system. The fact that the \deleted[]{control system}\added[]{DNN} without the current state feedback can still have a comparable improvement offers us an alternative offline method of improving tracking performance. It makes our approach more versatile, especially when computational resources are not sufficient to support real-time calculations during flights.

\begin{figure}[t]
    \centering
    \includegraphics[width=0.4\textwidth]{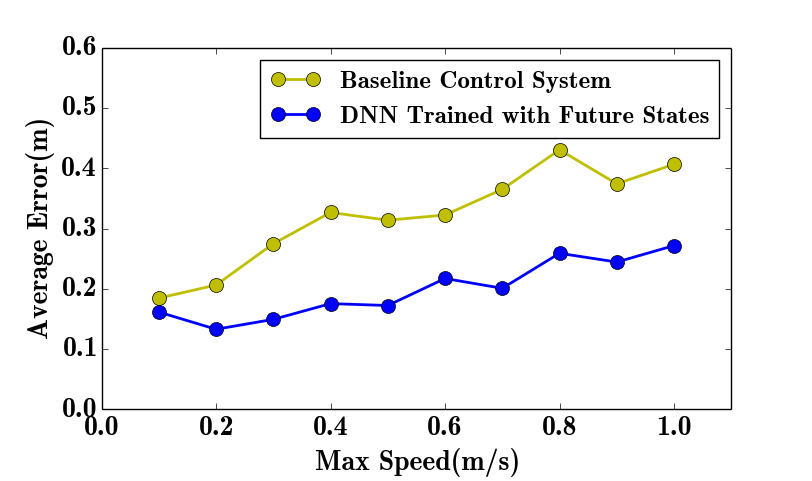}
    \caption{The average performance of the DNNs trained with two future desired states ($\Delta_1$ = 4, +0.6 s, and  $\Delta_2$ = 6, +0.9 s ahead of the current time) and the \deleted[]{original}\added[]{baseline} control on \deleted[]{a specific trajectory}\added[]{Trajectory 4 in Figure~\ref{fig:9drawn}} with different speeds. The trajectories with different speeds were obtained from the drawn trajectory by changing the velocity bound. For the same trajectory, DNNs are able to obtain consistent improvement in tracking performance over different speeds.}
    \label{fig:DNN_dspd}
\end{figure}

\begin{table}[bt]
\centering 
\caption{Tracking errors on one trajectory for three different control systems}

    \begin{tabular}{|p{0.13\textwidth} | p{0.07\textwidth} | p{0.1\textwidth} | p{0.07\textwidth}|} 
    \hline 
    & Baseline Controller & DNN without Feedback & DNN with Feedback  
    \\\hline 
    RMS Error, $E$ (m) & \textcolor{black}{0.360} & \textcolor{black}{0.232} & \textcolor{black}{0.144}   
    \\\hline
    Peak Error (m)& 0.605& 0.497& 0.356  
    \\\hline
    Improvement, $I$~(\%) & - & \textcolor{black}{35.6}&\textcolor{black}{59.9} 
    \\\hline 
\end{tabular}

\label{table:fbvsnfb} 
\end{table}


\subsection{DNN Tracking Performance on Arbitrary Trajectories}

\begin{figure}[t]
    \centering
    \includegraphics[width=0.4\textwidth]{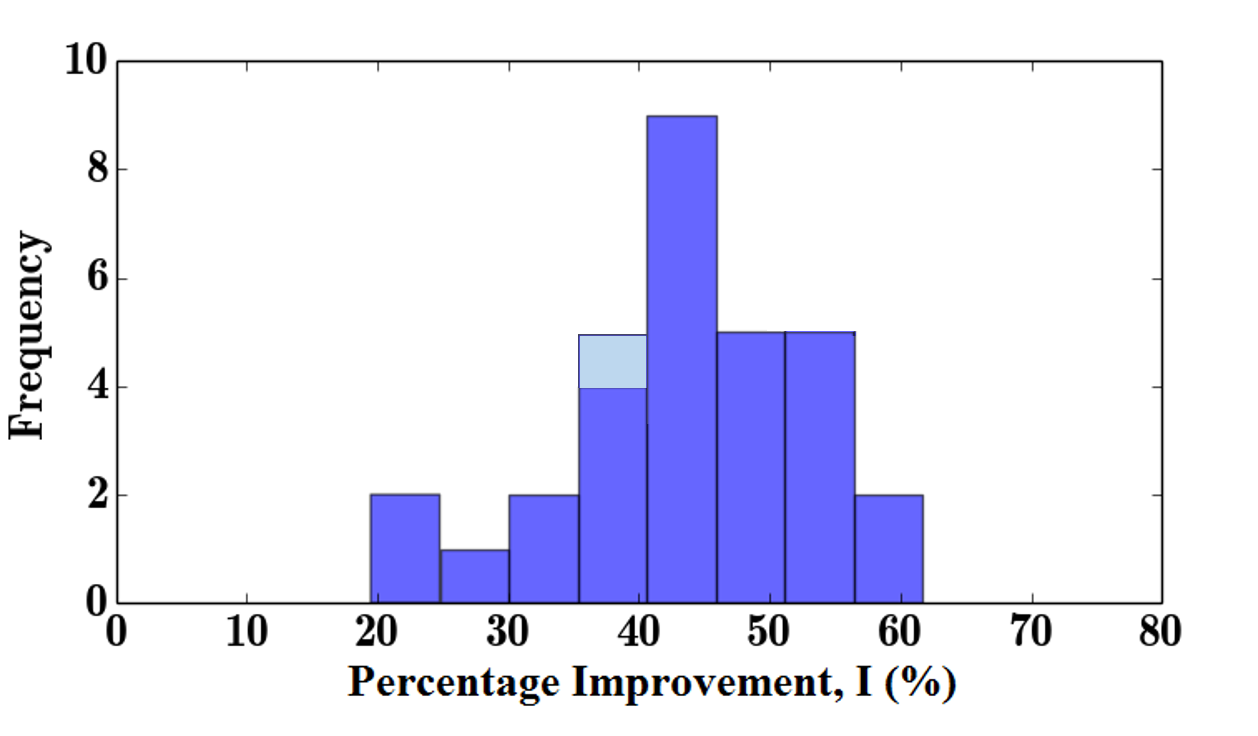}
    \caption{The improvements, \emph{I}, made by the proposed control system trained with two future desired states ($\Delta_1 = 4$, +0.6~s, and  $\Delta_2 = 6$, +0.9~s ahead of the current time) compared to the baseline control on 30 testing trajectories and a 50~s segment from the training trajectory. DNNs achieved \textcolor{black}{36\%} improvement on the training trajectory segment and it is represented by light-blue block on the graph. On average, 43\% improvement is obtained by the control system with DNNs.}
    \label{fig:DNN_hist}
\end{figure}

\begin{figure}[t]
    \centering
    \vspace{3mm}
    \includegraphics[width=0.35\textwidth]{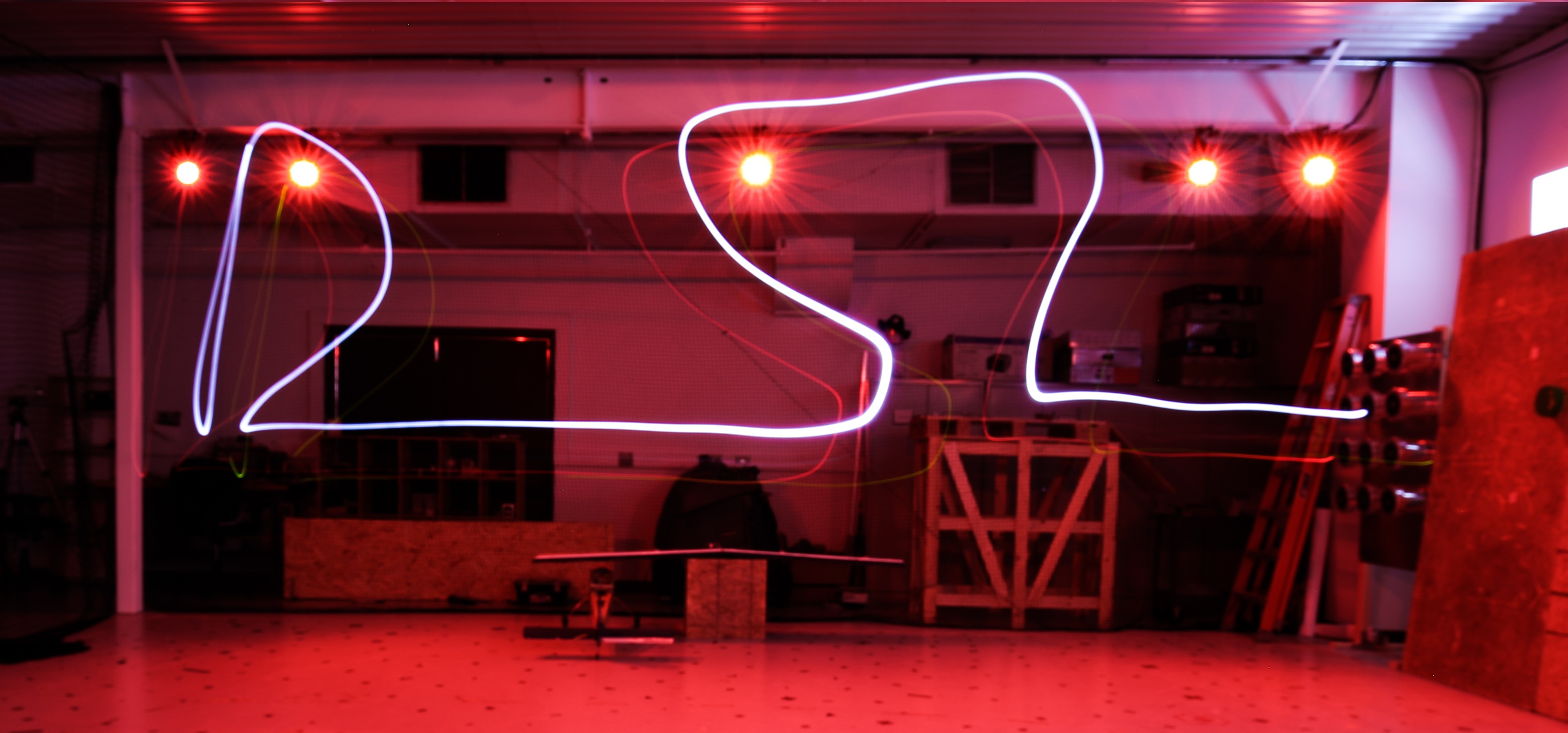}
    \caption{A long exposure image of a quadrotor following the letters ``DSL" with the aid of DNNs. The trajectory is drawn by a visitor. The DNNs take in 2 future desired states ($ \Delta_1 = 4$, +0.6 s, and $\Delta_2 = 6$, +0.9 s) and the current state feedback.}
    \label{fig:DNN_long_exp}
\end{figure}

To investigate generalizability of the trained DNNs on different trajectories, we evaluate the performance of the trained DNNs with future desired states and current state feedback on various trajectories. On a \added[]{50s }segment from the 3-D training trajectory (as discussed in Section \ref{subsec:data}), the DNNs outperform the baseline controller by \textcolor{black}{36\%}. 
The performance of the trained DNNs is also evaluated on unseen trajectories, including 30 different hand-drawn trajectories and one specific trajectory (Trajectory 4 on Figure \ref{fig:9drawn}) with different velocity profiles\footnote{The 30 drawn trajectories all have a maximum velocity of 0.6 m/s and a maximum acceleration of 2 m/s$^2$. For Trajectory 4, the speed is changed by scaling the time domain along the desired trajectory.}. In \deleted[]{Figure \ref{fig:DNN_bar} and }Figure \ref{fig:DNN_hist}, we show that the DNNs with future desired states and the current state feedback reduce the RMS tracking errors by 43\% on average over the 30 testing trajectories and training trajectory segment. Similar improvement is also obtained on one specific trajectory with different speeds as shown in Figure \ref{fig:DNN_dspd}. Therefore, the DNNs trained with future desired states are capable of reducing the tracking error for trajectories with various shapes and speeds by a large margin, demonstrating its generalizability on different unseen trajectories. Figure \ref{fig:DNN_long_exp} presents a long-exposure image of a quadrotor following the letters ``DSL" written by a visitor. Note that this is different from Trajectory 5 shown in Figure \ref{fig:9drawn} \added[]{since this is the write-up of another visitor}.

\addtolength{\textheight}{-3.4cm}   
\section{Conclusions}
\label{sec:conclusion}
In this paper, we have presented a DNN-based reference learning method\added[]{,} able to learn from flight data and improve trajectory tracking control in quadrotor systems. By introducing information about future desired states in training data, the DNNs were able to account for system delay and hidden dynamics as shown from the significant reduction in tracking error overall. The main advantages of this proposed approach shown from our experiments are that 1) this approach can be applied to various control systems with complex dynamics while ensuring stability of the systems; \deleted[]{and }2) it requires no prior knowledge of the system to train the DNNs, and the trained DNNs can be applied to any unseen trajectories without any adaptation process; and 3) with wise feature selection and sufficient DNN training, this approach can be computationally efficient with a very small model, while demonstrating good performance on general trajectories. We have shown these advantages through the implementation of an interactive ``fly-as-you-draw" application, illustrating that the proposed method was readily applicable to various real-world trajectory tracking tasks. However, the overall improvements of this method are still limited by training data for the DNNs. Intelligent choices of learning targets and effective neural network designs are potential extensions to enhance the trajectory tracking performance.  

\bibliographystyle{IEEEtran}
\bibliography{IEEEabrv,bibliography}

\begin{thebibliography}{10}
\providecommand{\url}[1]{#1}
\csname url@rmstyle\endcsname
\providecommand{\newblock}{\relax}
\providecommand{\bibinfo}[2]{#2}
\providecommand\BIBentrySTDinterwordspacing{\spaceskip=0pt\relax}
\providecommand\BIBentryALTinterwordstretchfactor{4}
\providecommand\BIBentryALTinterwordspacing{\spaceskip=\fontdimen2\font plus
\BIBentryALTinterwordstretchfactor\fontdimen3\font minus
  \fontdimen4\font\relax}
\providecommand\BIBforeignlanguage[2]{{%
\expandafter\ifx\csname l@#1\endcsname\relax
\typeout{** WARNING: IEEEtran.bst: No hyphenation pattern has been}%
\typeout{** loaded for the language `#1'. Using the pattern for}%
\typeout{** the default language instead.}%
\else
\language=\csname l@#1\endcsname
\fi
#2}}

\bibitem{drug_traffickers}
\BIBentryALTinterwordspacing
``{Central American Drug Compound Recon},'' October 2010. [Online]. Available:
  \url{https://www.aeryon.com/casestudies/centralamericadrug}
\BIBentrySTDinterwordspacing

\bibitem{drone_rescue}
\BIBentryALTinterwordspacing
C.~Gothner, ``{Deputies using drones as search-and-rescue tools},'' August
  2016. [Online]. Available:
  \url{http://www.channel3000.com/news/deputies-using-drones-as-searchandrescue-tools/41297542}
\BIBentrySTDinterwordspacing

\bibitem{drone_delivery}
\BIBentryALTinterwordspacing
S.~Shaw, ``{7-Eleven Teams with Flirtey for First Ever FAA-Approved Drone
  Delivery to Customer's Home},'' July 2016. [Online]. Available:
  \url{http://corp.7-eleven.com/news/07-22-2016-7-eleven-teams-with-flirtey-for-first-ever-faa-approved-drone-delivery-to-customer-s-home}
\BIBentrySTDinterwordspacing

\bibitem{michael2011cooperative}
N.~Michael, J.~Fink, and V.~Kumar, ``{Cooperative manipulation and
  transportation with aerial robots},'' \emph{Autonomous Robots}, vol.~30,
  no.~1, pp. 73--86, 2011.

\bibitem{5569026}
Q.~Lindsey, N.~Michael, D.~Mellinger, and V.~Kumar, ``{The grasp multiple
  micro-uav testbed},'' \emph{IEEE Robotics \& Automation Magazine}, vol.~17,
  no.~3, pp. 56--65, 2010.

\bibitem{Lupashin201441}
S.~Lupashin, M.~Hehn, M.~W. Mueller, A.~P. Schoellig, M.~Sherback, and
  R.~D'Andrea, ``{A platform for aerial robotics research and demonstration:
  The Flying Machine Arena},'' \emph{Mechatronics}, vol.~24, no.~1, pp. 41--54,
  2014.

\bibitem{i_2015_DL_models}
A.~Punjani and P.~Abbeel, ``{Deep learning helicopter dynamics models},'' in
  \emph{IEEE International Conference on Robotics and Automation (ICRA)}, 2015,
  pp. 3223--3230.

\bibitem{mohajerin2014modular}
N.~Mohajerin and S.~L. Waslander, ``{Modular deep Recurrent Neural Network:
  Application to quadrotors},'' in \emph{IEEE International Conference on
  Systems, Man, and Cybernetics (SMC)}, 2014, pp. 1374--1379.

\bibitem{a_2014_DIC}
M.~T. Frye and R.~S. Provence, ``{Direct Inverse Control using an Artificial
  Neural Network for the Autonomous Hover of a Helicopter},'' in \emph{IEEE
  International Conference on Systems, Man, and Cybernetics (SMC)}, 2014, pp.
  4121--4122.

\bibitem{Schoellig2012}
A.~P. Schoellig, F.~L. Mueller, and R.~D'Andrea, ``{Optimization-based
  iterative learning for precise quadrocopter trajectory tracking},''
  \emph{Autonomous Robots}, vol.~33, pp. 103--127, 2012.

\bibitem{mueller2012iterative}
F.~L. Mueller, A.~P. Schoellig, and R.~D'Andrea, ``{Iterative learning of
  feed-forward corrections for high-performance tracking},'' in \emph{IEEE/RSJ
  International Conference on Intelligent Robots and Systems}, 2012, pp.
  3276--3281.

\bibitem{2013_ILC_transfer}
M.~W. Michael~Hamer and R.~D'Andrea, ``{Knowledge Transfer for High-Performance
  Quadrocopter Maneuvers},'' in \emph{IEEE/RSJ International Conference on
  Intelligent Robots and Systems (IROS)}, 2013, pp. 1714--1719.

\bibitem{2014_surgical}
J.~Mahler, S.~Krishnan, M.~Laskey, S.~Sen, A.~Murali, B.~Kehoe, S.~Patil,
  J.~Wang, M.~Franklin, P.~Abbeel, \emph{et~al.}, ``{Learning accurate
  kinematic control of cable-driven surgical robots using data cleaning and
  gaussian process regression},'' in \emph{IEEE International Conference on
  Automation Science and Engineering (CASE)}, 2014, pp. 532--539.

\bibitem{lecun2015deep}
Y.~LeCun, Y.~Bengio, and G.~Hinton, ``{Deep learning},'' \emph{Nature}, vol.
  521, no. 7553, pp. 436--444, 2015.

\bibitem{kingma2014adam}
D.~Kingma and J.~Ba, ``{Adam: A method for stochastic optimization},''
  \emph{arXiv preprint, arXiv:1412.6980}, 2014.

\bibitem{2014-dropout}
N.~Srivastava, G.~Hinton, A.~Krizhevsky, I.~Sutskever, and R.~Salakhutdinov,
  ``{Dropout: A Simple Way to Prevent Neural Networks from Overfitting},''
  \emph{Journal of Machine Learning Research}, vol.~15, pp. 1929--1958, 2014.

\end{thebibliography}

\newpage

\end{document}